\documentclass{egpubl}
\usepackage{amsmath}
\usepackage{amssymb}
\usepackage{float}
 \JournalSubmission    
\usepackage[T1]{fontenc}
\usepackage{dfadobe}  
\usepackage{multirow}
\usepackage{xcolor}
\usepackage{cite}  
\BibtexOrBiblatex
\electronicVersion
\PrintedOrElectronic
\ifpdf \usepackage[pdftex]{graphicx} \pdfcompresslevel=9
\else \usepackage[dvips]{graphicx} \fi

\title[Row-Column Separated Attention Based Low-Light Image/Video Enhancement]%
      {Row-Column Separated Attention Based Low-Light Image/Video Enhancement}

\author[Dong et al.]{
{\parbox{\textwidth}{\centering Chengqi Dong$^{1}$,
        Zhiyuan Cao$^{2}$,
        Tuoshi Qi$^{1}$,
        Kexin Wu$^{1}$,
        Yixing Gao$^{1}$,
        and Fan Tang$^{3}$ }
        }
        \\
{\parbox{\textwidth}{\centering $^1$School of Artificial Intelligence, Jilin University, China\\
$^2$College of Software, Jilin University, China\\
         $^3$Institute of Computing Technology, Chinese Academy of Sciences, China}}
         }
 
\begin{document}


\maketitle
\begin{abstract}
U-Net structure is widely used for low-light image/video enhancement.
The enhanced images result in areas with large local noise and loss of more details without proper guidance for global information.
Attention mechanisms can better focus on and use global information. However, attention to images could significantly increase the number of parameters and computations.
We propose a Row-Column Separated Attention module (RCSA) inserted after an improved U-Net.
The RCSA module’s input is the mean and maximum of the row and column of the feature map, which utilizes global information to guide local information with fewer parameters. 
We propose two temporal loss functions to apply the method to low-light video enhancement and maintain temporal consistency. Extensive experiments on the LOL, MIT Adobe FiveK image, and SDSD video datasets demonstrate the effectiveness of our approach. The code is publicly available at https://github.com/cq-dong/URCSA.



\end{abstract}  
\section{Introduction}
\label{sec:intro}

Limited by realistic objective conditions, we encounter many low-light images and videos. 
A low-light environment leads to imaging degradations, including intense noise, low visibility, chromatic aberrations, and information loss~\cite{xu2022deep,lv2018mbllen,ko2017artifact}. Low-light image enhancement is proposed to improve the quality of images captured under low-light conditions. 
The main goal of the low-light image enhancement algorithm is to create visually pleasing images and provide more information than input images suitable for applications~\cite{teng2019remote,6512558}.
Low-light data enhancement methods can improve the aesthetic quality of low-light visual media and contribute to data pre-processing of visual tasks~\cite{lv2018mbllen,zhang2021learning}. 
The goal of image enhancement is to remove image noise, increase brightness, improve the structural similarity of images, improve visual perception, and reduce color difference.

With the development of deep learning, neural network-based methods~\cite{Chongyi,lv2018mbllen,zamir2020learning,tu2022maxim,fan2022half} have emerged. 
Through a well-designed network, many encoder and decoder models represented by the U-Net structure~\cite{zhang2021learning, Chi-Mao, Shangchen} can achieve better enhancement results.
However, since these methods perform multiple downsampling operations and only use convolution kernels to extract features, the enhanced results may lose many local details and lack global guidance. This means the enhancement effect worsens in low-light areas with high local noise. Currently, there are also global information guidance methods based on the transformer. 
Still, the transformer-based methods only accept relatively small patches of fixed sizes, which inevitably cause patch boundary artifacts when applied on larger images using cropping~\cite{tu2022maxim}.
Therefore, the size of the input image is limited, and the number of parameters is large, which means training and inference are more time-consuming. To address the fact that the computational complexity of ViT is quadratic with the resolution of input, LLFormer~\cite{wang2023ultra} proposed A-MSA to compute self-attention on the height and width axes across the channel dimension sequentially and employed depth-wise convolutions. STAR~\cite{zhang2021star} flattened the full-size image into a sequence of image blocks and then reduced the spatial resolution by Adaptive Average Pooling to make dimensionality reduction of the image blocks, which significantly reduced the complexity and computation of generating tokens. 
Other researchers have also conducted research on row and column attention, such as the RCANet~\cite{lu2022rcanet} paper used for semantic segmentation, which separately sets row and column attention operations on each channel and then merges them without cross-channel information exchange.
The Lanformer~\cite{han2022laneformer} paper used for lane detection uses the row and column feature vectors extracted by the network for attention operations and adds the row and column results as the output of the attention module, which cannot obtain pixel-level attention results. 
We propose row-column separated attention to extract global information with fewer parameters and faster training. 
Our RCSA module allows direct cross-channel information interaction, obtains pixel-level attention results, and achieves self-balancing between maximum and average attention.

\begin{figure}[!t]
    \centering
    \includegraphics[width=8cm]{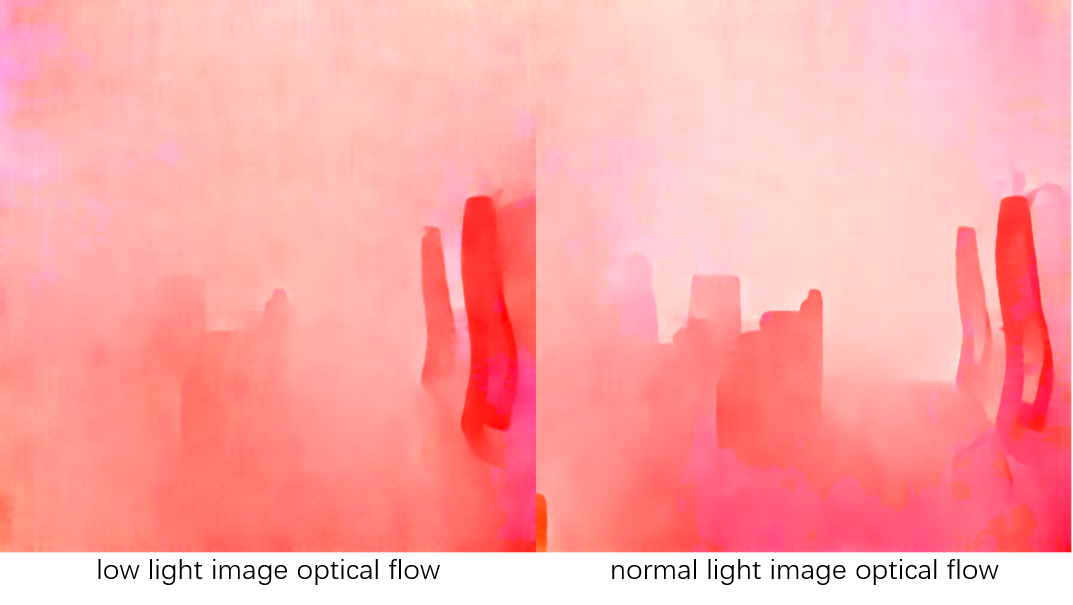}
    \caption{The light flow results of the low and normal light scenes are calculated by the GMflow~\cite{xu2022gmflow} model, in which the light flow of the dark scene is more blurred and missing many details.}
    \label{flow}
\end{figure}

Furthermore, we attempt to apply the method to video enhancement.
Nevertheless, directly using image enhancement methods on videos will lead to obvious flicker and jitter~\cite{xu2022deep}.
Therefore, the main challenge from image to video is achieving temporal stability of video frames while ensuring the quality of single-frame enhancement.
A common approach is to extend 2D image models to 3D video models, such as MBLLEN~\cite{lv2018mbllen}.
However, replacing the convolution kernel lacks global information and requires much computation. 
In addition to the 3D convolution model, some methods use optical flow information to enhance the video.
They use optical flow to describe the motion of objects in dynamic scenes, such as StableLLVE~\cite{zhang2021learning}.
However, such methods are affected by the accuracy of optical flow estimation. Specifically, calculating optical flow in low-light environments is more challenging, as shown in Fig.~\ref{flow}.
To ensure the stability of enhancement results, we propose two loss functions based on the characteristics of adjacent video frames. These functions consider the local differences and self-similarity of adjacent frames, respectively.

\begin{figure*}[htbp]
    \centering
    \includegraphics[width=18cm]{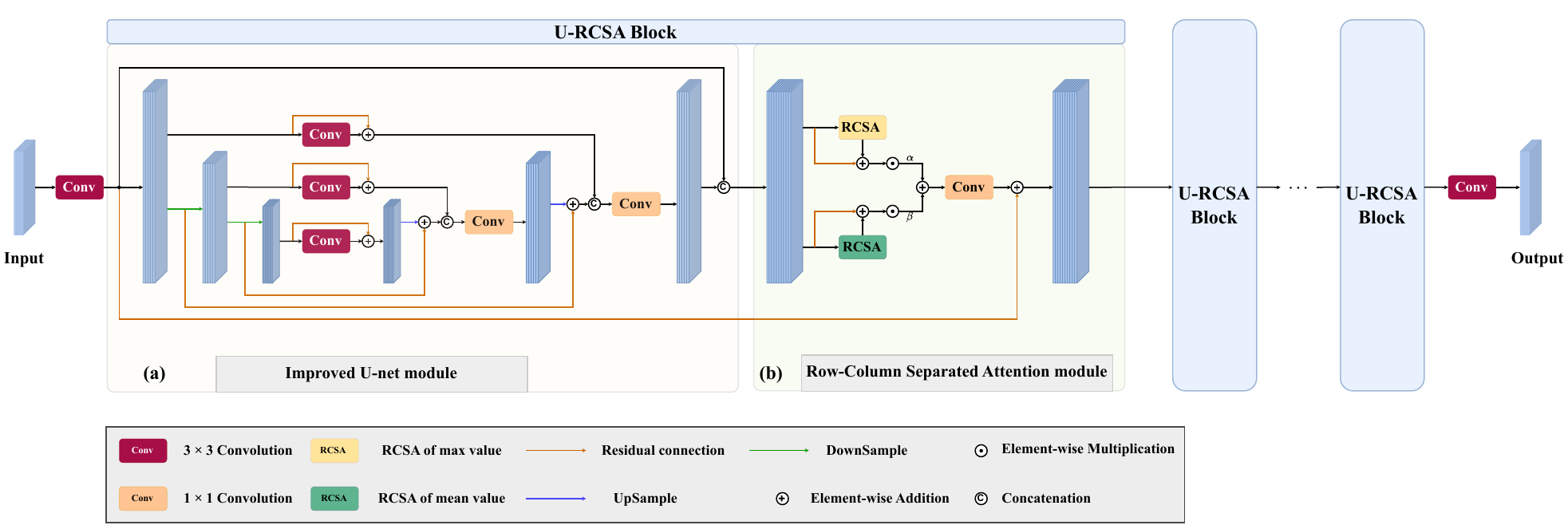}
    \caption{The framework of U-RCSANet. U-RCSANet consists of three U-RCSA blocks with the same parameters. Each U-RCSA block has an improved U-Net and a Row-Column Separated Attention module.}
    \label{framework}
\end{figure*}

In summary, considering that the existing methods lack global information guidance and are limited by the transformer methods, we innovatively propose a Row-Column Separated Attention (RCSA) module inserted after the improved U-Net. 
The RCSA module and improved U-Net form the U-RCSA block.
Furthermore, we design the U-RCSANet by stacking three U-RCSA blocks with the same parameters as shown in Fig.~\ref{framework}.
The improved U-Net is responsible for extracting local information and fusing shallow information with deep information in the framework. At the same time, the RCSA module has a small number of parameters and is not limited by image size, which is responsible for conducting guidance on global information. Our contributions are as follows.

\begin{itemize}
  \item We propose a novel U-RCSANet that leverages global context to guide local information while fusing shallow and deep features for low-light enhancement tasks.
  \item We propose an efficient and compact parameter RCSA module, which guides pixel-level attention results by learning global feature information of rows and columns.
  \item We propose two temporal loss functions to guide the model in learning the temporal stability information of the video. Extensive experiments on image and video datasets have proved that our method is more effective than the existing models.
\end{itemize}

\section{Related work}
\textbf{Low-light Image Enhancement.} 
Traditional low-light image enhancement methods, such as gamma correction~\cite{reinhard2010high}, histogram equalization~\cite{pizer1987adaptive}, and Retinex-based theory~\cite{rahman1996multi}, are usually based on the image prior knowledge or algorithms~\cite{Chi-Mao}. Reinhard et al.~\cite{reinhard2010high} edit the gamma curve of the image, detect the dark and light parts in the image signal, and increase the ratio between them, thereby improving the image contrast effect. Lee et al.~\cite{lee2013contrast} proposed a contrast enhancement algorithm based on the layered difference representation of 2D histograms, attempting to enhance image contrast by amplifying the gray-level differences between adjacent pixels. Based on the Retinex theory, Jobson et al.~\cite{jobson1997properties} proposed the best placement of the logarithmic function and Gaussian form to define a specific retinex called SSR to handle gray-world violations. Although these methods can significantly improve image brightness, the enhanced images have poor details and often suffer from severe color distortion~\cite{Chi-Mao}. Since LL-net~\cite{Kin} first utilized deep learning models for low-light enhancement, deep learning methods have become the dominant enhancement method~\cite{Chongyi}. Wei et al.~\cite{wei2018deep}, applying Retinex theory to deep learning, proposed RetinexNet, which decomposes images into reflectivity and illuminance and enhances illuminance to obtain normal light images. Lv et al.~\cite{lv2018mbllen} proposed a multi-branch network called MBLLEN that was designed to apply enhancement functions through multiple subnets and ultimately generate output images through multi-branch fusion to enhance and denoise low-light images simultaneously. The encoder-decoder structure represented by U-net has also been widely used, and the existing methods have achieved good results through well-designed network structures such as \cite{zhang2021learning, Chi-Mao, Shangchen}. Fan et al.~\cite{Chi-Mao} proposed HWMNet based on an improved hierarchical model M-Net+ and uses a half wavelet attention block on M-Net+ to enrich the features from the wavelet domain. Zhou et al.~\cite{zhou2022lednet} proposed LEDNet, consisting of an optical enhancement encoder and a deblurring decoder, and utilizes Filter Adaptive Skip Connections to eliminate ambiguity. 
Zhang’s StableLLVE~\cite{zhang2021learning} directly uses U-net network validation to increase the effectiveness of optical flow methods. 
However, the network of this structure often lacks the utilization of global information. Our proposed U-RCSA model emphasizes the guidance of global attention to local features, which are computed to obtain pixel-level attention results through row-column mean and maximum features. We combine deep and shallow information for interaction using subtle residual connection and channel concatenation. 

\textbf{Low-light video enhancement}. The biggest challenge facing the transition from low-light image to low-light video enhancement is achieving temporal stability of video frames while ensuring the quality of single-frame enhancement. This means avoiding problems such as flicker and jitter between video frames~\cite{xu2022deep}. Furthermore, obtaining high-quality paired low-light videos is also challenging, as it is necessary to ensure that dynamic scenes under different lighting conditions are identical. A simple approach is using 3D convolution to train continuous sequences of video frames, extending the image model to a video model. There are also methods to introduce optical flow information into the model's training process, and they use optical flow to describe the motion of objects in dynamic scenes through optical flow. Lai et al.~\cite{lai2018learning} also proposed a deep recurrent network with a ConvLSTM module that uses optical flow as a post-processing method for enforcing temporal consistency in a video. Zhang et al.~\cite{zhang2021learning} use optical flow prior to indicate potential motion from a single image and imitate adjacent frames of images by warping them with corresponding optical flow. They train the model in a twin mode and apply consistency between adjacent frames.

\section{Method}

As a supervised learning method,  given a paired dataset <$I_{in}, I_{gt}$>, the U-RCSA model enhances low-light image $I_{in}$ to a normal light image $I_{pred}$. The process can be formulated as:
\begin{equation}
    I_{pred} = \mathcal{F}\left(I_{in};\theta \right),
\end{equation}
where $\mathcal{F}$ represents our model U-RCSA with the trainable parameters $\theta$. 
We optimize the model by minimizing the error:
\begin{equation}
    \hat{\theta} = \underset{\theta }{argmin} \, \mathcal{L} \left( I_{pred},\; I_{gt} \right ),
\end{equation}
where $\mathcal{L}$ denotes the loss function. In the model, we propose the RCSA module with a few parameters in Sec.~\ref{c1} to focus on the global information. Furthermore, we improved the original U-Net, combined it with the RCSA module to form the U-RCSA block, and stacked it three times to form the U-RCSANet in Sec.~\ref{c2}. The network structure is shown in Fig.~\ref{framework}. Finally, to ensure the temporal consistency of the method applied to video, we proposed two loss functions in Sec.~\ref{c3}.

\subsection{Row-Column Separated Attention Module}
\label{c1}
In the existing research, popular attention modules for image processing, such as ViT~\cite{dosovitskiy2020image} and ViViT~\cite{arnab2021vivit}, are pixel-to-pixel attention by dividing the image into multiple small patches and then stretching the patches into vectors. ViT is to divide a picture into $16 \times 16$ patches of the same size and then stretch each $16 \times 16$ patch into a $1 \times 256$ vector. However, the number of parameters and the computational effort required for this processing are large. Limited by the image size, this processing is less suitable for image enhancement tasks. Inspired by separable convolution, maxpool and meanpool, we propose a novel attention module based on the row-column mean and maximum. We argue that the mean and maximum values of the rows and columns represent the row-column level information, which is local information relative to the entire feature map. However, the mean and maximum values of all rows and columns can represent the global information of the feature map. The feature map here refers to the input part of the RCSA module, that is, the feature map stitched by the channel from the input and output parts of the previous Improved U-Net module. By self-attention operation, we can fuse all the row-column feature information, thus obtaining the global information of the image.

\begin{figure}[htbp]
    \centering
    \includegraphics[width=9cm]{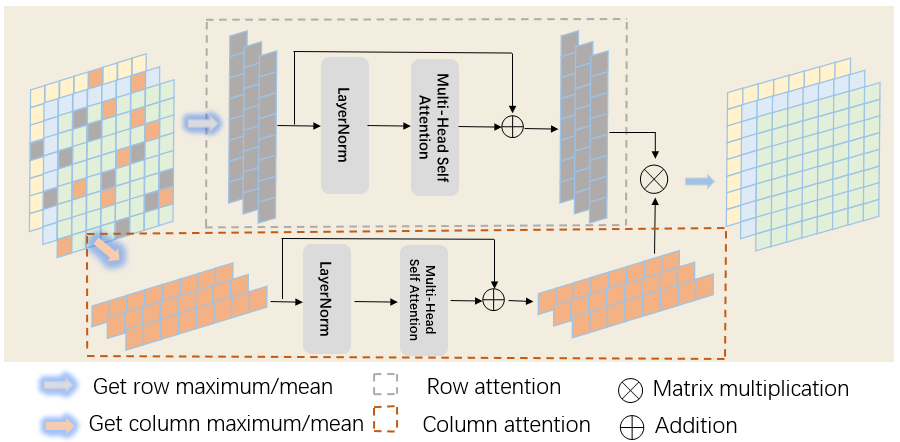}
    \caption{Row-Column Separated Attention module. Average and maximum attention can be obtained through column mean and maximum value.}
    \label{fig2}
\end{figure}

\begin{figure}[htbp]
    \centering
    \includegraphics[width=9cm]{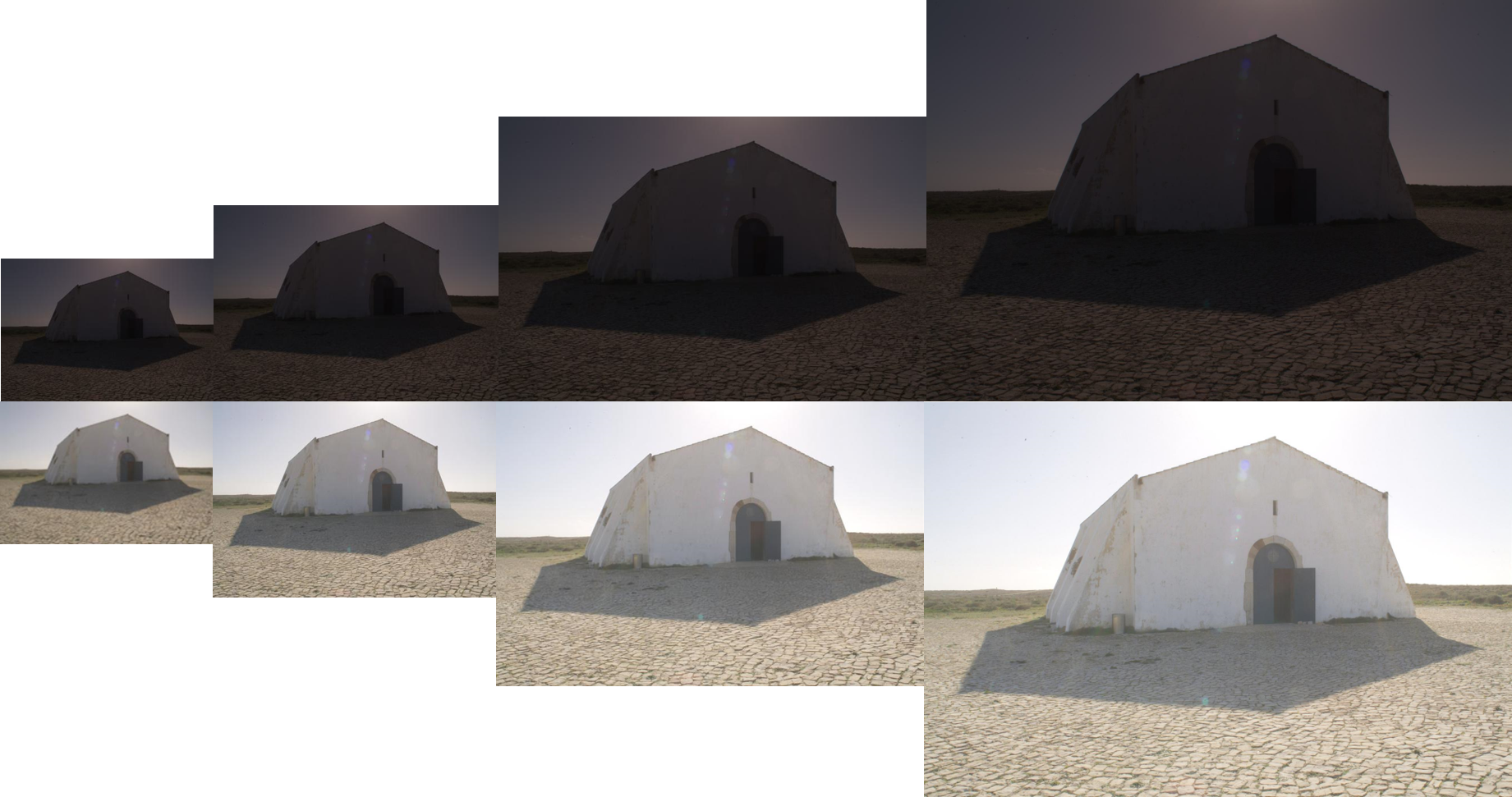}
    \caption{The results of dark light image enhancement for different sizes. From right to left, the image sizes are 1512$\times$1036, 768$\times$512, 512$\times$352, and 384$\times$256.}
    \label{fig0}
\end{figure}

Fig.~\ref{fig2} shows the specific details of the row-column separated attention module. First, we obtain $Fh_{avg}$ and $Fw_{avg}$ for the input feature map $F\,\in \mathbb{R}^{C\times H\times W}$ by row
and column respectively, and also obtain $Fh_{max}$ and $Fw_{max}$ by row and column respectively, where the features $Fh_{avg}$, $Fh_{max}\,\in \mathbb{R}^{C\times H\times 1}$ and $Fw_{avg}$, $Fw_{max}\,\in \mathbb{R}^{C\times 1\times W}$. We believe that they represent the average features of the feature map rows, the average features of the columns, the significant features of the rows, and the significant features of the columns. In this way, when processing the feature maps of the same number of channels, the amount of pixel information processed by our attention module is $(2/H+2/W)$ of the pixel attention of ViT~\cite{dosovitskiy2020image}, which leads to fewer parameters and faster calculation speed. Our attention module applies to images of any size, as shown in Fig.~\ref{fig0}. 

The input of mean and maximum can be calculated as follows:
\begin{equation}
\begin{array}{c}
	Fh_{avg}=\frac{1}{w}\sum_{i=1}^w{F\left( C,H,i \right)},\,\,\,\,Fh_{max}=\underset{i}{\overset{w}{\max}}\,F\left( C,H,i \right),\\
	Fw_{avg}=\frac{1}{h}\sum_{i=1}^h{F\left( C,i,W \right)},\,\,\,\,Fw_{max}=\underset{i}{\overset{h}{\max}}\,F\left( C,i,W \right).\\
\end{array}
\end{equation}
Then, the attention's output can be multiplied by the matrix to obtain a dimensional consistent output. This result is obtained by multiplying the corresponding row and column results, thus ensuring unique differences between pixels.

After completing the matrix multiplication, we obtain two attention results: mean attention and maximum attention. Since they represent the image's average and significant properties, we perform an adaptive weight balancing operation by setting a parameter for each channel separately. The final output can be calculated as:
\begin{equation}
	F_{out}=\boldsymbol{a}\cdot \varPhi \left( Fh_{avg} \right) \otimes \varPhi \left( Fw_{avg} \right)+\boldsymbol{b}\cdot \varPhi \left( Fh_{max} \right) \otimes \varPhi \left( Fw_{max} \right),
\end{equation}
where $\boldsymbol{a}=sigmoid\left( \boldsymbol{\lambda } \right)$, $ \boldsymbol{b}=1-sigmoid\left(\boldsymbol{\lambda } \right)$, $\otimes$ represents matrix multiplication, and $\varPhi \left( \boldsymbol{x} \right) =Softmax\left( \frac{QK^T}{\sqrt{d}} \right) V$  represents attention operation. 
$Q, K, V$ are the corresponding query, key, and value matrices in the attention mechanism, which are obtained by matrix multiplication of three learnable parameter matrices $W_{q}, W_{k}$ and $W_{v}$ with the input $x$, $\boldsymbol{\lambda }$ are adaptive weight parameters.

In order to prove the effectiveness of the RCSA module, we replace our RCSA module with the following modules, which are channel attention module (CAM)~\cite{Woo_2018_ECCV}, spatial attention module (SAM)~\cite{Woo_2018_ECCV}, SE attention block from SeNet~\cite{hu2018squeeze} and ECA attention module from ECANet~\cite{wang2020eca}. See Sec.~\ref{4.4} for a comparison of quantitative results.
\subsection{U-RCSANet}
\label{c2}
\textbf{Improved U-Net.}
As shown in Fig.~\ref{framework}, we perform two down-sampling operations to balance the computational effort while obtaining deep image feature information. After down-sampling, we perform an up-sampling process with a symmetric structure. Unlike U-Net, we do not directly concatenate the channel dimensions of the encoder and decoder~\cite{ronneberger2015u}. We perform residual and feature re-extraction operations for the encoder output, adding residual structures for each upsampling operation and performing additional feature extraction operations on the encoder results before performing channel cascading operations. After the concatenating operation, we recover the channels by 1 $\times$ 1 convolution layer. That is to say, we fuse shallow and deep features by concatenating the output results of the corresponding layers of the encoder and decoder in the channel dimension and extracting features again from the concatenation results through convolutional layers. Overall, we emphasize the importance of this residual and channel concatenation, which is beneficial in achieving the guidance of shallow information to deep information and can reduce the accumulation of errors caused by model extraction of deep features.

\noindent \textbf{U-RCSA block.}
The Improved U-Net is connected to the RCSA module as shown in Fig.~\ref{framework}. We concatenate the input and output of the Improved U-Net and use the result as the input of the RCSA module. This concatenation is essential and guides shallow information (before reconstruction by Improved U-Net) to deep information (after reconstruction by Improved U-Net), reducing the accumulation of noise introduced by the model reconstruction. The model achieves the interaction of deep and shallow information through subsequent feature extraction operations on the concatenated results, thereby providing guidance for deep information from shallow information. At the end of the attention module, we use 1 $\times$ 1 convolution to reduce the channel dimension and add a residual connection. Channel concatenation and residual connection between different modules are carefully designed, significantly improving the model's enhancement capabilities. To form U-RCSANet, we stack three U-RCSA blocks whose parameters are the same. The dimensions of channels are elevated and reduced through convolution at both ends of the network.

\subsection{Objectives}
\label{c3}
We have designed two loss functions for different contexts: the image quality loss function and the video loss function.

\noindent \textbf{Image loss function.} We adopted the loss function commonly used for low light enhancement and drew on the loss setting scheme in SMNet~\cite{lin2023smnet}, combining multiple functions to ensure the model can achieve better results.
\begin{itemize}

\item SSIM loss~\cite{wang2004image}: The loss is a constraint on structural similarity and helps to produce high-quality results.
\begin{equation}
\mathcal{L} _s=1-\left( \frac{2\mu _x\mu _y+C_1}{\mu _{x}^{2}+\mu _{y}^{2}+C_1}\cdot \frac{2\sigma _{xy}+C_2}{\sigma _{x}^{2}+\sigma _{y}^{2}+C_2} \right), 
\end{equation}
where $\mu _x$ and $\mu _y$ are the mean values of the input and output image, and present variances, $\sigma _{xy}$ is the covariance, and C1 and C2 are the constants that keep the denominator from being zero.
\item VGG loss~\cite{johnson2016perceptual}: We use the output features of the VGG~\cite{simonyan2014very} network to calculate the perceptual loss. The loss helps the model to fuse the semantic information of the image with the structural information better.
\begin{equation}
\mathcal{L} _p=\left\| \phi _{ij}\left( I_{pred} \right) -\phi _{ij}\left( I_{gt} \right) \right\| _1, 
\end{equation}
where $\phi _{ij}$ describes feature maps computing with $j$-th convolution layer of $i$-th block in VGG-16 network.
\item Smooth loss~\cite{girshick2015fast}: Compared to traditional L1 loss or L2 loss, Smooth L1 loss has a smoother curve, which can reduce the training instability caused by noise or outliers and better balance stability and accuracy.
\begin{equation}
\mathcal{L} _{\mathrm{sl}1}=\frac{1}{H\cdot W}\sum_{i=1}^H{\sum_{j=1}^W{\left\{ \begin{array}{c}
	\frac{1}{2}\left( I_{pred}\left( i,j \right) -I_{gt}\left( i,j \right) \right) ^2,\\
	if\,\,\left| I_{pred}\left( i,j \right) -I_{gt}\left( i,j \right) \right|<1\\
	\\
	\,\left| I_{pred}\left( i,j \right) -I_{gt}\left( i,j \right) \right|-0.5\,\,,\\
	if\,\,\left| I_{pred}\left( i,j \right) -I_{gt}\left( i,j \right) \right|\geq 1
\end{array} \right..}}
\end{equation}

\item TV loss~\cite{simonyan2014very}: Noisy images are often accompanied by high variance~\cite{paszke2017automatic}, so reducing the total variation between adjacent pixels can help achieve cleaner output.
\begin{equation}
    \mathcal{L} _{tv}=\sum_{i=1}^H{\sum_{j=1}^W{\sqrt{\left( I\left( i,j \right) -I\left( i+1,j \right) \right) \left( I\left( i,j \right) -I\left( i,j+1 \right) \right)}}},
\end{equation}
where $I\left( i,j \right)$ represents the pixel at the position $\left( i,j \right)$ in the enhanced image.
\item MSE loss $\mathcal{L} _{MSE}$: We use MSE loss to regularize the model and improve its robustness.
\end{itemize}

\noindent \textbf{Video loss function.} To ensure the temporal stability of enhanced video, we propose the following video loss functions.
\begin{itemize}

\item Neighboring-frame significant difference loss. Considering that the light and shadow flow in the local area of adjacent frames of dynamic video is not obvious and the object motion amplitude is small, we design the loss function:
    \begin{equation}
    \mathcal{L}_{dif}=\left| \mathrm{pool}\left( bright_k \right) -\mathrm{pool}\left( bright_{k+1} \right) \right|,
    \end{equation}
where the \(pool(x)\) represents the maximum pooling operation with a pooling core size of 2 and a step size of 2 and \(bright_k\), \(bright_{k+1}\) represent the brightness characteristics of the enhanced neighboring frames respectively. Its calculation formula is $bright_k=I_{k}^{r}*0.299+I_{k}^{g}*0.587+I_{k}^{b}*0.114$, where $I_{k}^{r}, I_{k}^{g}, I_{k}^{b}$ represents the R, G, and B channels of the k-th frame image, respectively. The local flicker phenomenon can be avoided by minimizing the local brightness difference between neighboring frames.
\item Self-similarity loss. Inspired by StableLLVE~\cite{zhang2021learning}, we believe that the neighboring frame differences of the normal light video should be consistent with the neighboring frame differences of the enhanced video. This is obvious because it allows the model to learn more about the potential temporal stability information between the two video frames of the input. Therefore, we propose the following loss function.
\begin{equation}
\mathcal{L}_{self}=\left\| \left( pred_{k+1}-pred_k \right) -\left( gt_{k+1}-gt_k \right) \right\| ^2,
\end{equation}
where \(gt_k\), \(gt_{k+1}\) are the adjacent frames of normal light video, \(pred_k\), \(pred_{k+1}\) are the adjacent frames of the corresponding enhanced video.
\end{itemize}
\noindent \textbf{Training for image}:
\begin{equation}
\begin{aligned}
\begin{cases}
	\mathcal{L} _{stage1}=\mathcal{L} _p+\mathcal{L} _s+\mathcal{L} _{sl1}+0.001*\mathcal{L} _{tv}, \\
	\mathcal{L} _{stage2}=\mathcal{L} _p+\mathcal{L} _s+\mathcal{L} _{sl1}+\mathcal{L} _{MSE},\\
\end{cases}\,\,
\end{aligned}
\end{equation}
where $\mathcal{L} _{stage1}$ is trained for the first stage to focus on improving the quality of image enhancement, and $\mathcal{L} _{stage2}$ is trained for the second stage to regularize the model and enhance its robustness.

\noindent  \textbf{Training for video}:
\begin{equation}
\mathcal{L} _{video}=\alpha\cdot \mathcal{L} _{dif}+\beta\cdot \mathcal{L} _{self}+\mathcal{L} _{stage1},
\end{equation}
where $\alpha$ and $\beta$ are the balancing coefficients.
We set $\alpha=2$ and $\beta=2$.

\section{Experiment}
To verify the model's validity, we trained and evaluated the model on public image and video datasets, respectively. In Sec.~\ref{4.1}, we introduced the datasets and experimental setup. In Secs.~\ref{4.2} and \ref{4.3}, we introduced the quantitative and qualitative experimental results, respectively. Finally, we presented the ablation study of the model in Sec.~\ref{4.4}.

\subsection{Experimental setup and datasets}
\label{4.1}
\textbf{Datasets.} The LOL dataset consists of 500 pairs of low and normal light images, of which 485 pairs are used for training and 15 pairs for testing~\cite{wei2018deep}. MIT Adobe FiveK~\cite{fivek} is a collection of 5000 RAW format photos taken by a group of photographers using DSLR cameras, covering a wide range of scenes, themes, and lighting conditions. 
SDSD~\cite{wang2021seeing} is the high-quality paired low-light video and normal-light video data collected by professional equipment. It includes indoor data, SDSD indoor and outdoor data SDSD outdoor. SDSD indoors has 58 pairs of video data for training and 12 pairs of data for testing, while SDSD outdoors has 67 pairs of video data for training and 13 pairs of video data for testing. 

\begin{table*}[htbp]
\centering
\caption{Evaluation results of different methods on the LOL dataset. Note that we cannot find an open-source model for experimentation in the blank part of the table.}
\label{table1}
\begin{tabular}{ll|cccccc}
\hline
Methods&Venue                  & PSNR$\uparrow$                     & SSIM$\uparrow$                     & LPIPS$\downarrow$                     & NIQE$\downarrow$                            & RMSE$\downarrow$         & DeltaE$\downarrow$       \\ \hline
LIME~\cite{guo2016lime} &TIP 2016              & 16.76                     & 0.56                    &           -              &              -                      &        -       &      -          \\
Retinex~\cite{wei2018deep}  &CVPR 2018              & 16.77                     & 0.56                     & 0.474                     & 8.865                              & 10.09       & 60.70          \\
MBLLEN~\cite{lv2018mbllen} &BMVC 2018           & 17.56 & 0.73 & 0.174 & \textbf{3.585} & 9.99        & 56.55          \\
ENlightenGAN~\cite{jiang2021enlightengan} &TIP 2021              & 17.48                     & 0.65                      & 0.173                     & 4.684              &   10.07           &62.72                \\
MIRNet~\cite{zamir2020learning}  &ECCV 2020             & 24.14                     & 0.83                     & 0.131                     & 4.201                              & 8.78        & 28.02          \\
TPET~\cite{cui2022tpet} &EAAI 2022              & 23.98                     & 0.84                    & 0.122                     &                 -                   & 24.46      &         -       \\
MAXIM~\cite{tu2022maxim}  &CVPR 2022            & 23.43                     & 0.86                     & 0.099                     & 5.330                               & 9.11          & 33.06          \\
HWMNet~\cite{fan2022half} &ICIP 2022              & 24.24                     & 0.85                     & 0.114                     & 5.141                              & 8.75          & 29.59          \\
Ours       &  & \textbf{24.77}            & \textbf{0.86}            & \textbf{0.097}            & 4.638                              & \textbf{8.64} & \textbf{25.49} \\ \hline
\end{tabular}
\end{table*}
\textbf{Experimental setup.} The source code provided in this article is implemented using the PyTorch framework and trained using the Adam optimizer, with the parameter betas set as the default value. Betas is the coefficient used to calculate the running average of the gradient and its square, with default values of $\beta _1=0.9$ and $\beta _2=0.999$. In order to enhance data while reducing training costs, we randomly cropped regions of 128×128 in the LOL image for training. Correspondingly, the regions cropped randomly in the SDSD dataset were 128×240. All test data obey the dataset corresponding to the paper settings. We set the initial learning rate to 5e-4, decreasing by 1.2 times every 50 epochs for LOL, every 20 epochs for MIT Adobe FiveK, and every 2 epochs for SDSD. LOL trained a total of 600 epochs, and  MIT Adobe FiveK trained a total of 100 epochs. Because the SDSD dataset is too large, we only train 10 epochs. We adopted the staged loss function, which improved both metrics, in particular, PSNR by 2\%. The metrics of other models in the following experiments are preferentially obtained from published papers. Otherwise, we use the source code provided for experiments.
\subsection{Quantitative evaluation}
\label{4.2}
\textbf{LOL dataset.} We compared the U-RCSANet with the latest model on the LOL image dataset. We chose the two most mainstream indicators, PSNR and SSIM. Higher PSNR generally represents less noise, and higher SSIM represents more similar structural properties. Our model exceeds all other models in these two mainstream indicators. In addition, we have also selected LPIPS~\cite{zhang2018unreasonable}, NIQE~\cite{saad2012blind}, RMSE, and DeltaE~\cite{robertson1977cie} to measure the effectiveness of the model comprehensively.
LPIPS~\cite{zhang2018unreasonable} stands for Learned Perceptual Image Patch Similarity, used to measure the similarity between two images based on their perceptual features. A high LPIPS score indicates that the two images being compared are not similar in their perceptual features. NIQE~\cite{saad2012blind} means Natural Image Quality Evaluator, which measures the quality of a distorted or compressed image. A higher score indicates poorer quality and implies that the distorted or compressed image being evaluated has a lower quality than the original, undistorted image. RMSE represents Root Mean Square Error, and a higher RMSE score means that the two images being compared are more dissimilar. DeltaE~\cite{robertson1977cie} is a metric used in color science to measure the difference between two colors. A lower DeltaE score indicates a smaller color difference between the two colors. The results are shown in Table~\ref{table1}. It shows that our model is significantly superior to other models in multiple indicators. Our model parameters are less than the latest models HWMNet~\cite{fan2022half} and MAXIM~\cite{tu2022maxim} in 2022, and only 2\% of the HWMNet model, but our quantitative indicators are all ahead of both.

\begin{figure}[!t]
    \centering
    \includegraphics[width=9cm]{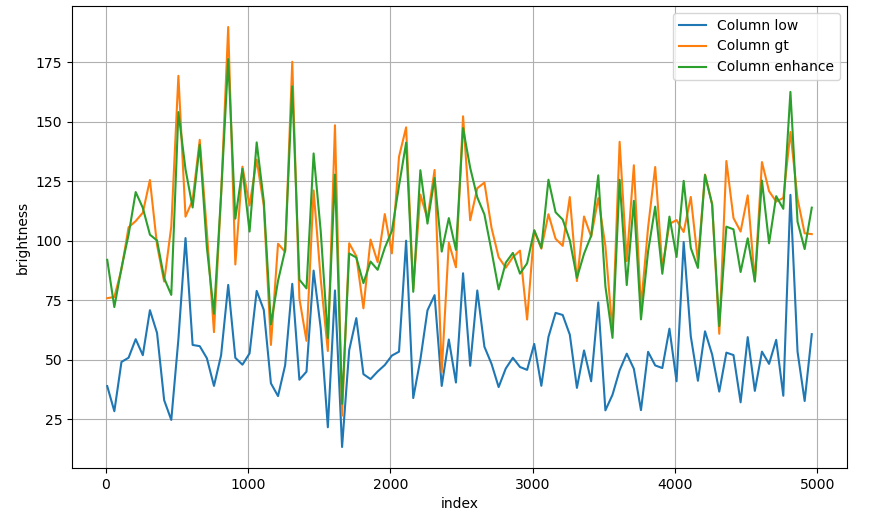}
    \caption{MIT FiveK dataset low light image, normal light image and enhanced image brightness visualization results, it can be found that the brightness of the enhanced image is consistent with the brightness of the normal light image through the degree of fluctuation of the three curves can be sent to achieve the model can be adaptive to the dark light image brightness enhancement.}
    \label{figbright}
\end{figure}

\textbf{MIT Adobe FiveK dataset.}
To verify the generality of the method, we trained on the MIT Abode FiveK dataset of 5000 paired datasets. Although the MIT dataset has a wide range of lighting conditions, our model performs well in brightness adaptive enhancement, as shown in Fig.~\ref{figbright}. 

We followed \cite{wang2019underexposed} for our experiments and chose the version of Student C as the ground truth. We followed the specific data partitioning method~\cite{wang2023ultra}, with 500 images used for testing and related evaluation metrics results also derived from LLFormer~\cite{wang2023ultra}. We selected classical models as well as the most up-to-date ones. Among them, Uformer~\cite{9878729}, Restormer~\cite{Zamir_2022_CVPR}, and LLformer are all transformer-based models which optimize the transformer mechanism in terms of time complexity. In particular, the LLformer model successively uses row-column attention to reduce the transformer complexity to linear order. The experimental results are shown in Table~\ref{table999}. Compared with the latest model LLformer, although the PSNR of our model does not achieve first place, there is only a 1‰ difference. 
However, our SSIM and LPIPS metrics are in the lead. More importantly, the number of parameters in our model is only about 12\% of that.

\textbf{SDSD dataset.} We also compared the model with the latest model on the video dataset. For image enhancement quality, we still chose PSNR and SSIM metrics. The results are shown in Table~\ref{table2}. The PSNR of our model is 2\% higher than that of DP3DF~\cite{xu2022deep} and 8\% higher than that of LSEC~\cite{aghajanzadeh2022long}. We selected AB, MABD, TSSIM, and TPSNR common metrics for temporal stability. AB metric to measure the average brightness variance between the enhanced video and the ground truth. This metric reflects when the video has unexpected brightness changes or flickers~\cite{lv2018mbllen}. Mean absolute brightness differences (MABD) can be viewed as a general level of time derivatives of brightness value on each pixel location~\cite{9010274}. The Mean squared error between the MABD vector of the enhanced video and the MABD vector of the ground live is the actual MABD index. TPSNR and TSSIM apply the PSNR and SSIM formulas to videos. The calculation process is as follows: first, the PSNR and SSIM indicators between adjacent frames of the video are solved, and then the average of the PSNR indicators of the video is calculated as TPSNR. Therefore, the higher the value, the better. The average of the SSIM indicators of the video is calculated as TSSIM, and the higher the value, the better. The result is shown in Table~\ref{table3}. The first two models are the image-based and optical flow-based video methods, and the last three are the ablation experiments of video loss of our model. It can be found that our temporal stability is greater than that of image-based methods. Although our method is not as good as the method based on optical flow in terms of quantitative indicators comparison, our model has also achieved good results without allowing for calculating optical flow outside the boundary.
\begin{table}[htbp]
\small
\centering
\caption{Evaluation results of different models on MIT Adobe FiveK dataset.}
\label{table999}
\begin{tabular}{ll|ccc}
\hline
Methods           & Venue & PSNR↑           & SSIM↑          & LPIPS↓         \\ \hline
LIME~\cite{guo2016lime} &TIP 2016        & 13.303          & 0.750          & 0.132          \\
Retinex~\cite{wei2018deep}  &CVPR 2018   & 12.515          & 0.671          & 0.254          \\
DSLR~\cite{9264763}       &TMM 2021    & 20.244          & 0.829          & 0.153          \\
Z\_DCE++~\cite{9369102}   & TPAMI 2022    & 14.611          & 0.406          & 0.231          \\
ELGAN~\cite{jiang2021enlightengan} &TIP 2021    & 17.905          & 0.836          & 0.143          \\
Uformer~\cite{9878729}    & CVPR 2022   & 21.917          & 0.871          & 0.085          \\
Restormer~\cite{Zamir_2022_CVPR}  & CVPR 2022   & 24.923          & 0.911          & 0.058          \\
LLFormer~\cite{wang2023ultra}   & AAAI 2023   & \textbf{25.753} & 0.923          & 0.045          \\
Ours       &    & 25.724          & \textbf{0.930} & \textbf{0.041} \\ \hline
\end{tabular}
\end{table}

\begin{table}[htbp]
\small
\centering
\caption{Evaluation results of different models in SDSD dataset.}
\label{table2}
\begin{tabular}{ll|cc}
\hline
Methods             & Venue      & PSNR$\uparrow$ & SSIM$\uparrow$ \\ \hline
DeepUPE~\cite{wang2019underexposed}             & CVPR 2019 & 21.82 & 0.68  \\
MBLLEN~\cite{lv2018mbllen}              & BMVC 2018 & 21.79 & 0.65  \\
SMID~\cite{chen2019seeing}                & ICCV 2019 & 24.09 & 0.69  \\
SMOID~\cite{jiang2019learning}               & ICCV 2019 & 23.45 & 0.69  \\
SDSD~\cite{wang2021seeing}                & CVPR 2021 & 24.92 & 0.73  \\
StableLLVE~\cite{zhang2021learning}              & CVPR 2021 & 22.85 & 0.79  \\
LSEC~\cite{aghajanzadeh2022long}   & CVPR 2022 & 25.06 & 0.84  \\
DP3DF~\cite{xu2022deep} & AAAI 2022 & 26.46 & 0.81  \\
\multicolumn{2}{l|}{Ours}       & \textbf{27.01} & \textbf{0.84}  \\ \hline
\end{tabular}
\end{table}

\begin{table}[h]
\small
\centering
\caption{Temporal Stability Comparison of Models on SDSD Datasets. Red is the best, blue is the second, and green is the third.}
\label{table3}
\begin{tabular}{l|llll}
\hline
& AB$\downarrow$ & MABD$\downarrow$  & TPSNR$\uparrow$   & TSSIM$\uparrow$                            \\ \hline
MBLLEN         & {\color[HTML]{3166FF} 0.17}          & 52.36                                 & 28.72                                 & 0.84                                 \\
StableLLVE         & {\color[HTML]{FE0000} \textbf{0.11}} & {\color[HTML]{FE0000} \textbf{11.01}} & {\color[HTML]{FE0000} \textbf{35.02}} & {\color[HTML]{FE0000} \textbf{0.91}} \\ \hline
w/o\,\,$\mathcal{L} _{dif}\,\,and\,\,\mathcal{L} _{self}$   & 0.63                                 & {\color[HTML]{32CB00} 16.73}          & 32.47                                 &{\color[HTML]{32CB00} 0.878}                                \\
w/o\,\,$\mathcal{L} _{self}$    & {\color[HTML]{32CB00} 0.24}          & 16.83                                 & {\color[HTML]{32CB00} 32.88}                                 & 0.871        \\
w/o\,\,$\mathcal{L} _{dif}$ & 0.63           & 17.35          & 32.00          & 0.873              \\
Ours            & 0.26                                 & {\color[HTML]{3166FF} 14.54}          & {\color[HTML]{3166FF} 33.02}          & {\color[HTML]{3166FF} 0.886}         \\ \hline
\end{tabular}
\end{table}

\begin{figure*}[htbp]
    \centering
    \includegraphics[width=18cm]{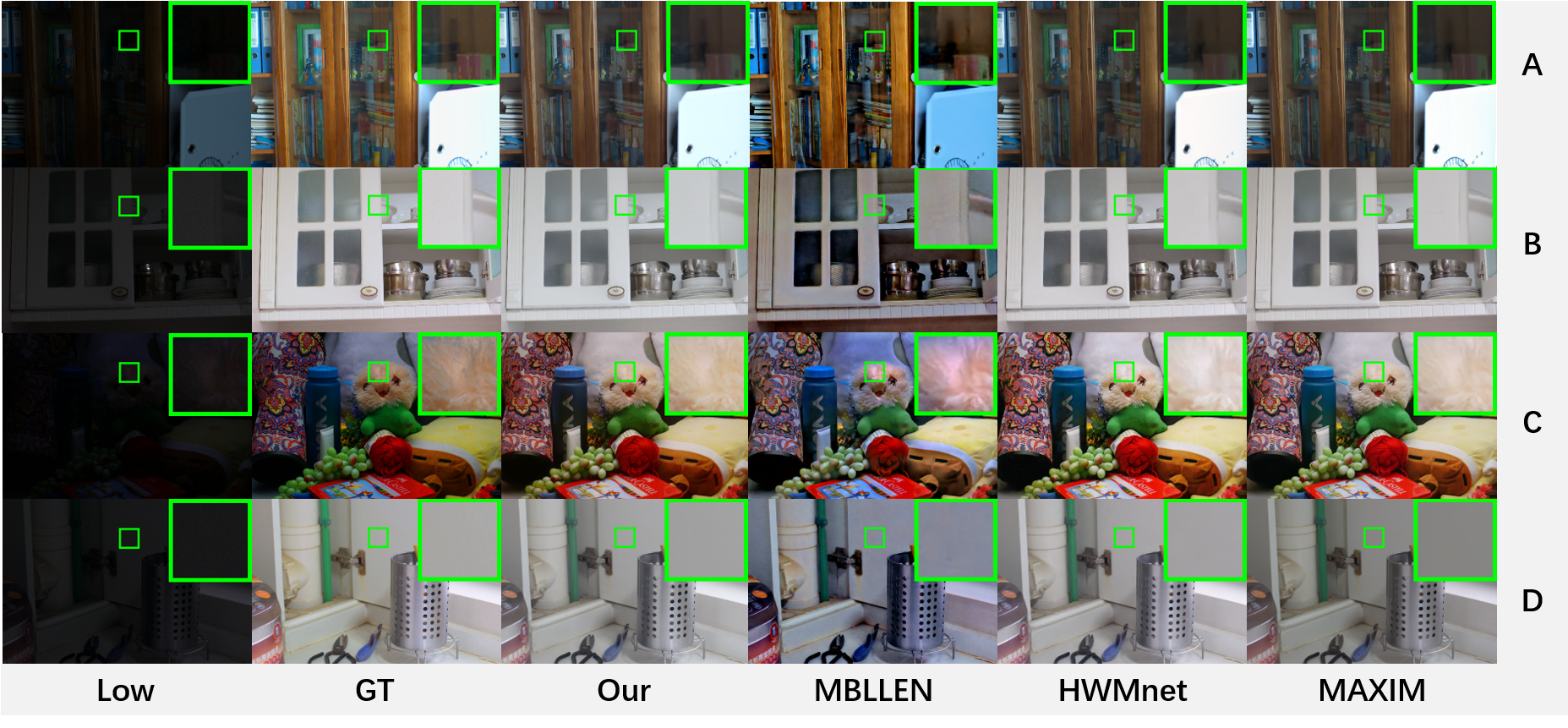}
    \caption{Comparison of enhancement results of different models in LOL dataset. Here, we choose HWMNet, MAXIM, and classical methods to compare the overall details.}
    \label{fig3}
\end{figure*}

\begin{table*}[htbp]
\centering
\caption{Quantitative metrics comparison of different attention modules. Our RCSA module achieved the best results. Although our model's FPS(frames per second) index is slightly lower than some models, it can be found that the PSNR of other attention module enhancement results is far from the number of our model, and the noise effect has not been minimized as much as possible.}
\label{table4}
\begin{tabular}{l|llllllll}
\hline
& PSNR$\uparrow$  & SSIM$\uparrow$  & LPIPS$\downarrow$    & FSIM$\uparrow$    & NIQE$\downarrow$    & RMSE$\downarrow$  &DeltaE$\downarrow$        & FPS$\uparrow$       \\ \hline
No attention & 19.768          & 0.827          & 0.129          & 0.949          & 4.698          & 10.02         & 54.63          & \textbf{13.01}          \\

CAM          & 23.079          & 0.853          & 0.107          & 0.964          & 4.689          & 9.25          & 32.91          &9.11          \\
SAM          & 20.096          & 0.836          & 0.128          & 0.950          & 4.753          & 9.73          & 43.00          & 12.38         \\
SE block     & 23.685          & 0.854          & 0.115          & 0.964          & 4.728          & 8.86          & 30.78          & 12.73          \\
ECA block    & 23.062          & 0.850          & 0.113          & 0.964          & 4.657          & 9.31          & 42.79          & 12.75         \\
RCSA(ours)   & \textbf{24.767} & \textbf{0.864} & \textbf{0.097} & \textbf{0.966} & \textbf{4.638} & \textbf{8.64} & \textbf{25.49} & 11.70 \\ \hline
\end{tabular}
\end{table*}

\subsection{Qualitative evaluation}
\label{4.3}
\begin{figure*}[htbp]
    \centering
    \includegraphics[width=18cm]{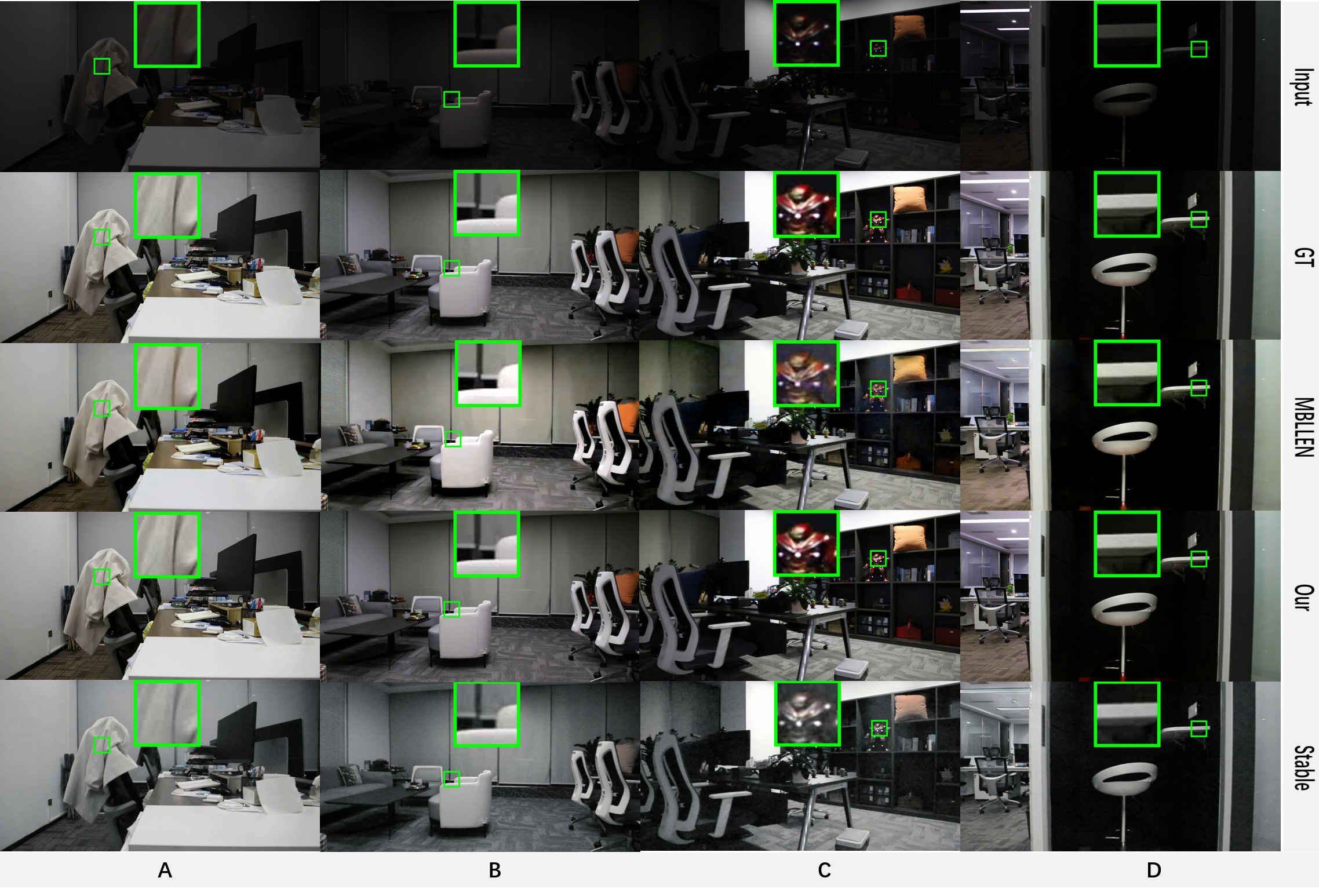}
    \caption{Comparison of enhancement results of different models in the SDSD dataset. Here, we choose MBLLEN and StableLLVE to compare the overall details. Our model can better restore complex details even in very dark areas than other models.}
    \label{fig4}
\end{figure*}

\begin{figure}[htbp]
    \centering
    \includegraphics[width=8cm]{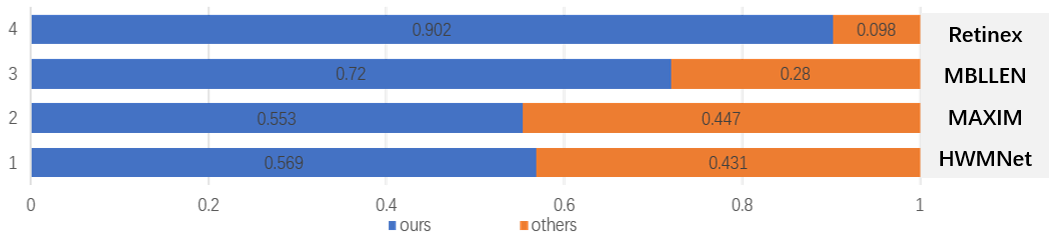}
    \caption{User study. Comparing our model with the two latest models in 2022 and a classical model on LOL dataset, our model has a higher selection rate.}
    \label{fig5}
\end{figure}

\begin{figure}[htbp]
    \centering
    \includegraphics[width=8cm]{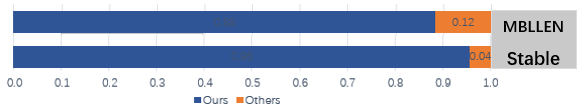}
    \caption{User study. Comparing our model with StableLLVE and MBLLEN on the SDSD video dataset, our model has a higher selection rate.}
    \label{fig6}
\end{figure}

We show qualitative results on two datasets. Fig.~\ref{fig3} shows a qualitative comparison of U-RCSANet with other models on the LOL test set. Generally, with the guidance of global information from the RCSA module, the U-RCSANet can better focus on the global brightness and restore the true colors and details. At the same time, MBLLEN has dark brightness in pictures B and C and distorted colors in A, C, and D. 
The four MAXIM images are all dark, and HWMNet also has a dark problem in pictures A and B, especially the hair color distortion in the enlarged area in picture C. We composited the test videos after enhancing all the images in the SDSD test dataset. The enhancement results can be seen in the additional materials. Fig.~\ref{fig4} shows the enhancement results on the model SDSD dataset. Through comparison, it can be found that there is always a greater difference between the color of the enhancement results and the ground truth in MBLLEN, and the overall enhancement results of the StableLLVE model are grayer and darker, while our model has better results. According to C in Fig.~\ref{fig4}, our model can better restore complex details even in very dark areas, while enhanced details by MBLLEN are more blurred and noisy.
Fig.~\ref{fig5} shows the user preference study. We conducted the user study using the LOL image enhancement results from the test set. Two choices were provided to the users each time. The results are from the 41 user preference choices. All of our models have higher selection rates. We also tested the statistical significance of the users' propensities based on the experimental data. We made the null hypothesis that the probability of our selected model is greater than 50\%, 60\%, 70\%, and 90\% compared to the other four models, MAXIM, HWMNet, MBLLEN, and Retinex, respectively. 
All the null hypotheses cannot be rejected at a significance level $\alpha$ of 0.05. Fig.~\ref{fig6} shows the comparison of video enhancement results of our model with the StableLLVE~\cite{zhang2021learning} model and the MBLLEN~\cite{lv2018mbllen} model. We collected the average selection results of 20 users for 16 videos of SDSD indoor and outdoor data. Although the StableLLVE model outperforms our model in terms of quantitative indicators of temporal consistency, user feedback indicates that neither model has significant flicker issues. Due to the poor image quality enhancement results of the StableLLVE model, our model has a higher selection rate. Although the image quality enhancement results of MBLLEN are better than those of StableLLVE, our model is also more popular due to the serious noise flicker problem of MBLLEN.

\begin{figure}[htbp]
    \centering
    \includegraphics[width=8cm]{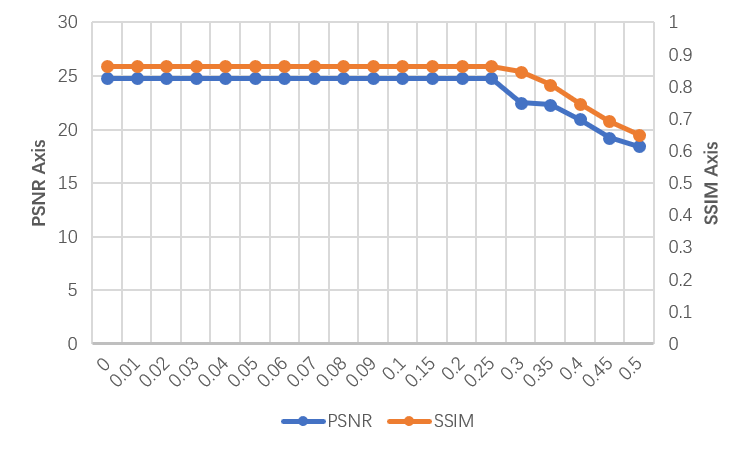}
    \caption{Quantitative noise experiments. We added Gaussian noise after image normalization, and the PSNR and SSIM were evaluated as above. When the standard deviation of the noise exceeds 0.25, the model's performance begins to degrade significantly.}
    \label{noise1}
\end{figure}

\begin{figure*}[htbp]
    \centering
    \includegraphics[width=18cm]{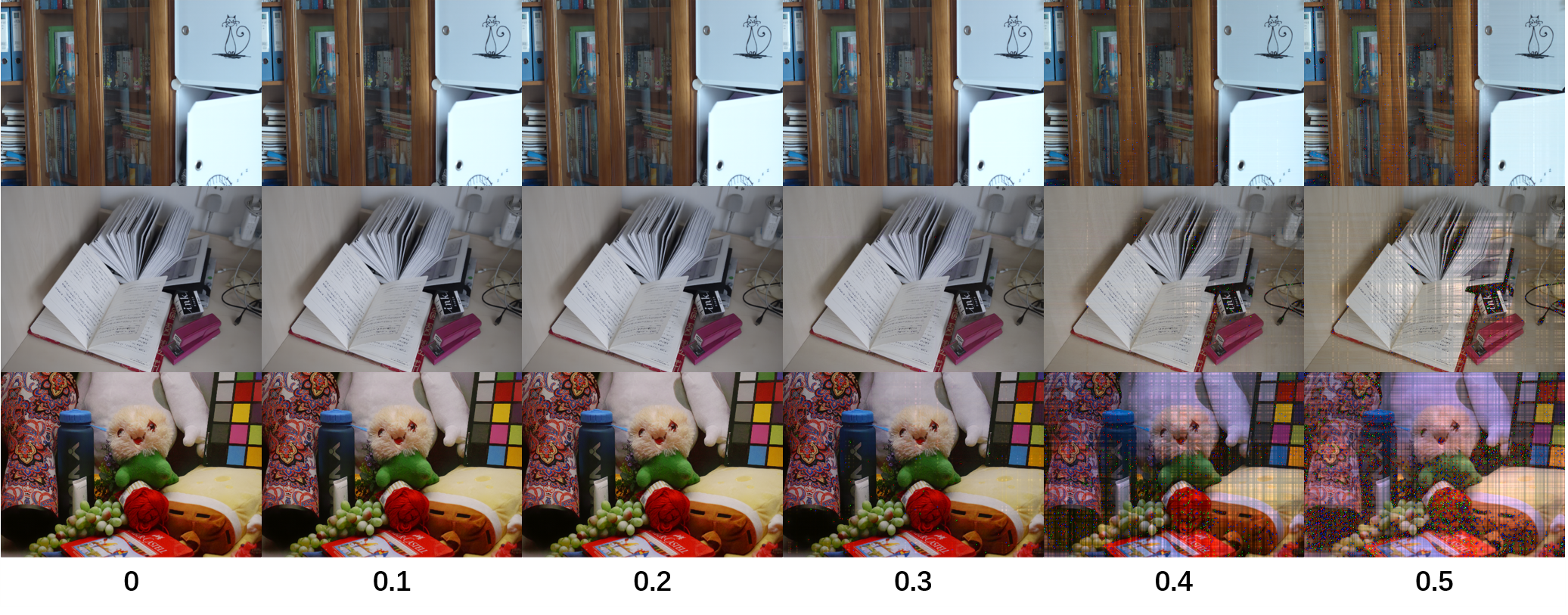}
    \caption{Qualitative noise experiments. We added Gaussian noise after normalizing the image. This figure shows the test results with a standard deviation of 0 to 0.5. The quality of the results begins to suffer when the standard deviation exceeds 0.3.}
    \label{noise2}
\end{figure*}

\noindent In addition, to verify the robustness of the model to noise, we conducted noise experiments on 15 LOL test set images. First, we normalize the image pixels by dividing them by 255 to normalize them to the [0,1] interval. Then we add Gaussian noise $\mathcal{N} \left( \mu,\sigma ^2 \right), \mu =0,\sigma \in \left[ 0,0.5 \right]$ to the normalized image data, and input the noisy image into the pre-trained model for testing. The experimental results can be seen in Fig.~\ref{noise1} and Fig.~\ref{noise2}, where Fig.~\ref{noise1} is a quantitative display and Fig.~\ref{noise2} is a qualitative display. \\

\subsection{Ablation study}
\label{4.4}

\begin{table}[htbp]
\centering
\caption{Model ablation study on LOL dataset.  `Initial U-Net with RCSA' represents the Three U-RCSA block's U-Net module without adding residual connection and convolution reprocessing. `The three RCSA-U blocks' are the model RCSA-U, which represents swapping the order of the improved Unet and RCSA modules in the original U-RCSA.}
\label{table6}
\begin{tabular}{l|ll}
\hline
Methods & PSNR$\uparrow$&SSIM$\uparrow$ \\ \hline
One U-RCSA block                  & 22.683                    & 0.846                     \\
Two U-RCSA blocks                 & 23.468                    & 0.855                     \\
Three U-RCSA blocks                & \textbf{24.767 }           & \textbf{0.864}                   \\
Three RCSA-U blocks                & 23.040                    &0.848                              \\
Three U-RCSA\_nomax blocks                & 23.164          & 0.852                 \\
Three U-RCSA\_noavg blocks                & 23.218           & 0.852                   \\
Four U-RCSA blocks                 & 23.865                    & 0.846     \\
Initial U-Net with RCSA &23.160 &0.852\\
Without RCSA module                      & 19.768                    & 0.827              \\
\hline
\end{tabular}
\end{table}
\noindent \textbf{Comparison of the effects of other lightweight attention modules and RCSA attention modules}: to verify the effectiveness and computational efficiency of the RCSA module, we replaced the RCSA module with CAM, SPA, SE block~\cite{hu2018squeeze}, and ECA block~\cite{wang2020eca}. These five modules are all lightweight, plug-and-play attention modules. We use the same method to train and test the LOL dataset. The FPS value is the average number of inference frames per second obtained by running 15 LOL test images on an RTX A5000 graphics card 200 times. As shown in Table~\ref{table4}, after adding any attention module, the PSNR and SSIM indicators were greatly improved, and our RCSA module achieved the best results, proving the effectiveness of the RCSA module compared to other lightweight attention modules. \\
\noindent \textbf{Effectiveness of U-RCSA block}: U-RCSA block is the main component of U-RCSANet. We explored the function of this module from the following two aspects. The experimental results were obtained in the LOL dataset and displayed in Table~\ref{table6}. 1) In the aspect of the number of U-RCSA block stacks. Quantitative results are shown in Table~\ref{table6}. The experimental results show that when the number of U-RCSA block stacks gradually increases from one to four, the PSNR and SSIM increase first and then decrease, achieving the best results when the number of stacks is three. Because our modules are not simply stacked, it can be seen from Fig.~\ref{framework} that there is a cross-block residual connection, and the results can be better enhanced through the information interaction of multiple blocks. 2) Whether to improve the U-net. The U-net of the U-RCSA block is our carefully designed and improved structure. Compared with the original U-net structure, the indicators of the improved structure significantly rose because we have added shallow information to the deep layer for the original U-net. The residual connection of information avoids the noise amplification caused by over-extracting information. It adds a feature convolution operation for the jump connection between the encoder and decoder to realize the feature conversion between coders better.

\begin{table}[htbp]
\caption{Comparison of parameter quantities and quantitative measures between the model in this article and the five most advanced models published in 2022 on the LOL dataset. Note: HWMNet, MAXIM, SNR, UNIE and LLFormer all contain their attention modules. In particular, LLformer is a Transformer-based model that reduces the runtime to linear complexity using the row-column attention mechanism. Our model has only one-eighth of its parameters, but the enhancement is superior to it.}
\label{table:table5}
\small
\begin{tabular}{ll|ccc}
\hline
Methods  & Venue   & Param           & PSNR  $\uparrow$         & SSIM  $\uparrow$         \\ \hline
Ours   &        & 11.9Mb          & \textbf{24.77} & \textbf{0.864} \\
Ours w/o RCSA   &        & \textbf{11.5Mb} & 19.77          & 0.830           \\
HWMNet~\cite{fan2022half} &   ICIP 2022     & 762Mb           & 24.24          & 0.852          \\
MAXIM~\cite{tu2022maxim}  &  CVPR 2022      & 14.1Mb          & 23.43          & 0.863          \\
SNR~\cite{xu2022snr}    &  CVPR 2022      & 152.8Mb         & 24.61          & 0.842          \\
UNIE~\cite{jin2022unsupervised}   & ECCV 2022 & 512Mb           & 21.52          & 0.760           \\
LLformer~\cite{wang2023ultra}  & AAAI 2023 & 282Mb           & 23.65          & 0.816           \\   \hline
\end{tabular}
\end{table}


\noindent \textbf{Effectiveness of the RCSA module}: We conducted the following experiments using the mean-attention and maximum-attention modules of the RCSA structure. 
When the RCSA module is not used, the two indicators drop significantly because the network lacks the extraction and utilization of global information after removing RCSA and cannot guide local areas due to large local noise. When we only use the mean or maximum value, the effect of the model also decreases because the mean and maximum values of the row and column represent different feature information of the row and column. Therefore, the adaptive accumulation of the mean and maximum matrices can avoid the excessive loss of information downsampling while utilizing different features as much as possible. 
Table~\ref{table:table5} compares our model with the five state-of-the-art models with attention operations in 2022 regarding parameter quantities and quantitative metrics on the LOL dataset. Compared with other attention-related models in 2022, our model is smaller and has the best effect. After adding the RCSA module, the number of our parameters only increased by 3\%. The model's size increased by 0.4Mb, but the PSNR and SSIM improved by 25\% and 4\%, respectively.\\

\noindent \textbf{Comparison of the network structures}: 
In order to verify the effectiveness of U-RCSA, a combination of UNet local convolution and RCSA global attention, we try to use the transformer structure and adjust the order of RCSA modules for comparison experiments. The results are shown in Table~\ref{table:table8}. 
Uformer, Restormer, and LLformer are the UNet-based network models that replace the local convolution module with the transformer module. 
Restormer achieves linear complexity by computing self-attention in the channel dimension through the multi-Dconv head `transposed' attention (MDTA) module. 
LLFormer significantly reduces linear complexity by the axis-based multi-head self-attention and cross-layer attention fusion block. 
In Table~\ref{table:table8}, U-Restormer is to substitute the RCSA module with Restormer's MDTA with linear complexity. U-RWKV is to replace the RCSA module with the RWKV~\cite{peng2023rwkv} model, where the RWKV model is a linear attention model that combines RNN and Transformer. The modified structure model follows the same training method on the LOL dataset. 
\begin{table}[htbp]
\centering
\caption{Comparison of various transformer-based structures. }
\label{table:table8}
\begin{tabular}{l|cccc}
\hline
Methods & PSNR $\uparrow$& SSIM $\uparrow$ & LPIPS $\downarrow$ \\ \hline
Uformer~\cite{Wang_2022_CVPR}   &18.547& 0.721 & 0.321\\
Restormer~\cite{zamir2022restormer}  &22.365& 0.816 & 0.141 \\
LLformer~\cite{wang2023ultra} &23.649& 0.816 & 0.169 \\ 
Retinexformer\cite{cai2023retinexformer}  & \textbf{25.160}& 0.845 & 0.131 &  \\ \hline
U-RWKV  &23.051& 0.855 & \textbf{0.095} \\
U-Restormer &23.900& 0.856 & 0.101 \\ \hline
RCSA-U  &23.040& 0.848 & 0.109\\
U-RCSA(Ours) &24.767& \textbf{0.864} & 0.097 \\ \hline
\end{tabular}
\end{table}

\begin{figure}[htbp]
    \centering
    \includegraphics[width=8cm]{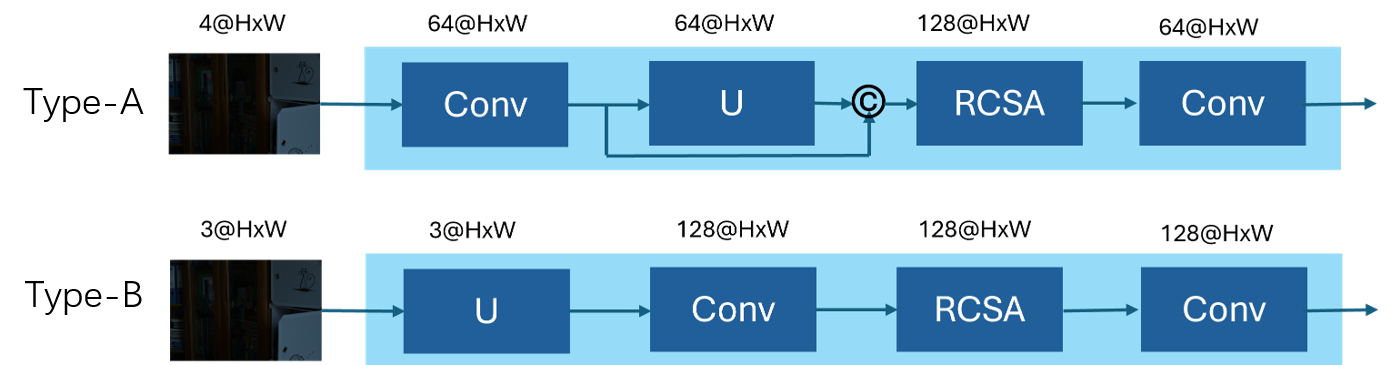}
    \caption{Two ways of fusion RCSA with Uformer. The blue area is a basic block. If it is a cascade model, there will be three blocks with shared parameters. The © represents channel concatenation, the $x$@H$\times$W above each component represents the output dimension of the component, and U represents the Unet or Uformer models. This combination is closest to our model. Type-A's input image has four channels, which means a brightness channel is added to the image.}
    \label{fig:typeab}
\end{figure}

In addition, we tested the ability of RCSA on Unet and Uformer and attempted two fusion structures, as shown in Fig. \ref{fig:typeab}. Type-A is closer to our model structure, and Type-B preserves the original structure of Uformer and Unet. After combining the RCSA module with Uformer and Unet, the results of both methods were improved, as shown in Table~\ref{table:result}, and the fusion method closer to the model in this paper had better results, proving the RCSA module's role. However, we found that Uformer and Unet may not be suitable for the cascade structure of this paper, and the effect of the Type-A fusion method was reduced after cascading.
\begin{table}[htbp]
\centering
\caption{Uformer/Unet combined with RCSA and cascade experiments. Cascade means the same three-time parameter sharing as in the article.}
\label{table:result}
\begin{tabular}{l|cccc}
\cline{1-3}
Methods  & PSNR $\uparrow$   & SSIM $\uparrow$   \\ \cline{1-3}
Type-A Unet with RCSA (cascade)     & 23.160 & 0.852 &  &  \\
Type-A Unet with RCSA              & 23.483 & 0.855 &  &  \\
Type-B Unet with RCSA (cascade)  & 21.873 & 0.829 &  &  \\
Type-B Unet with RCSA              & 22.370 & 0.845 &  &  \\
Type-A Uformer with RCSA (cascade)  & 24.254 & 0.859 &  &  \\
Type-A Uformer with RCSA           & 24.470 & 0.859 &  &  \\
Type-B Uformer with RCSA (cascade)  & 23.621 & 0.857 &  &  \\
Type-B Uformer with RCSA           & 22.668 & 0.860 &  &  \\ \cline{1-3}
\end{tabular}
\end{table}
\section{Limitation}
\begin{figure}[H]
    \centering
    \includegraphics[width=8cm]{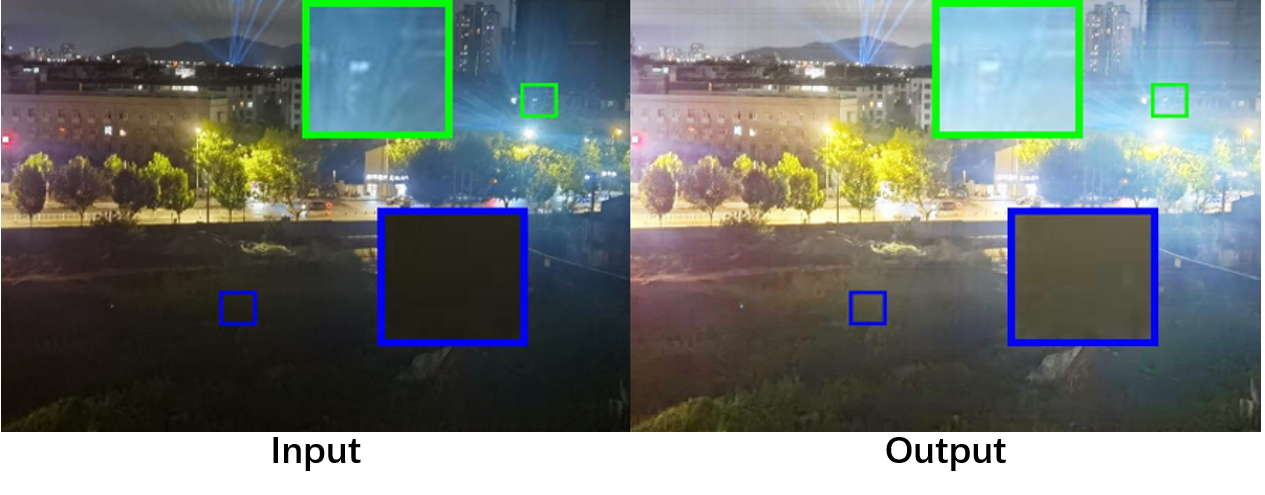}
    \caption{Failure cases in extreme environments.}
    \label{figbad}
\end{figure}

Although numerous experiments have shown that our method can achieve good results in most cases, when the input image is too dark or noisy, the enhancement effect of our model will deteriorate.
Fig.~\ref{figbad} shows a city night scene image we collected. It can be observed that the light source in the image is mixed, and some areas are too bright and too dark, as well as a large amount of noise. Moreover, due to the long shooting distance, the image itself is also relatively blurry. The green box in the figure has a moderate enhancement effect due to the influence of scattered light sources, while the blue box in the figure has a poor enhancement effect due to the area being too dark and the overall noise being high. In addition, although our model does not perform simple brightness enhancement on low-light images, it learns how to enhance low-light images into normal-light images that conform to human vision and achieve success in low-light image enhancement under different lighting conditions and scenes. Achieved good results, as shown in Fig.~\ref{figbright}. However, we still found that when the input is a normal light image, our model still has a certain degree of enhancement, reducing human vision's aesthetics. This will also be an improvement direction in the future, such as by adding similar training data or relevant constraints.

\section{Discussion}
The results in Table~\ref{table3} indicate that our temporal stability metrics exhibit a performance that lags behind those derived from optical flow techniques. 
At the same time, our model only relies on two adjacent frames of images. However, our proposed loss function still facilitates the model in learning the time stability information between video frames. Therefore, the $\mathcal{L}_{dif}$ and $\mathcal{L}_{self}$ loss functions are an idea and attempt to constrain the consistency between video frames. They can also be used in optical flow-based methods to help the model better constrain the inter-frame loss. Among them, the loss $\mathcal{L}_{dif}$ focuses on the constraints on the temporal consistency of the video itself, which is a universal temporal constraint. Imagine that if the input video frame has temporal inconsistency, there could be significant local brightness differences leading to a significant loss of $\mathcal{L}_{dif}$, which will force the model to minimize the local brightness differences of the enhancement results as much as possible to achieve better temporal consistency and make the model more robust to the data. The loss  $\mathcal{L}_{self}$ focuses on ensuring that the enhanced video is as close to the real video as possible, allowing the model to learn the temporal consistency of the training video's characteristics. In addition, our model only constrains the temporal consistency of video frames from the perspective of loss. A better approach would be to fuse attention and feature information between different frames, which will also be our future direction of effort.

\begin{table}[htbp]
\centering
\caption{Test results of different models on the LSRW dataset. The first three models were tested after they had been trained on LSRW. The following three models were directly tested on the LSRW test after they had already been trained on the LOL dataset.}
\label{GeneralizationResults}
\small
\begin{tabular}{ll|ccc}
\hline
Methods    & Venue      & PSNR $\uparrow$          & SSIM $\uparrow$         &NIQE $\downarrow$ \\ \hline
Retinex~\cite{wei2018deep}  & CVPR 2018 & 15.48          & 0.347       &  10.31 \\
KinD~\cite{zhang2021beyond}     & IJCV 2021 & 16.41          & 0.476    &  11.13    \\
URetinex~\cite{wu2022uretinex} & CVPR 2022 & 18.10          & 0.514      &  10.76  \\ \hline
SMNet~\cite{lin2023smnet}     & TMM 2023 & 17.59          & \textbf{0.521} & 3.97 \\
LLFormer~\cite{wang2023ultra} & AAAI 2023 & \textbf{18.13} & 0.519          &  \textbf{3.73}\\
Ours     &           & 17.97          & 0.513      &  3.86  \\ \hline
\end{tabular}
\end{table}
To verify the model's generalization ability, we conducted the following generalization test experiments on LSRW dataset~\cite{hai2023r2rnet}, and the results are shown in Table~\ref{GeneralizationResults}. SMNet, LLFormer, and ours obtained the results of pre-trained models from the LOL dataset tested directly on LSRW without fine-tuning. Retinex~\cite{wei2018deep}, KinD~\cite{zhang2021beyond}, and URetinex~\cite{wu2022uretinex} got the test results after training on the LSRW dataset. Although our model did not take the lead in the generalization experiments, the gap with the highest score was very small and similar to the test results of the pre-trained model. This proved to be a good generalization effect of our model.

\section{Conclusion}
We have modified the U-Net architecture to achieve a better fusion of deep and shallow information and proposed an RCSA module with a small number of parameters. Our method of taking row-column information as the basic unit of attention input can make good use of global information, and pixel-level attention can be obtained through the fusion of row-column attention results. The RCSA module can be easily migrated to other models as a lightweight module. Meanwhile, we set the loss functions for the features of adjacent frames, which guarantee the temporal stability of video frames. The results of a large number of comparative experiments on image and video datasets demonstrate the effectiveness of our model.
\section*{Data availability statement}

The LOL dataset that supports the findings of this study is available in Google Drive at \url{https://drive.google.com/file/d/157bjO1_cFuSd0HWDUuAmcHRJDVyWpOxB/view}, reference number~\cite{wei2018deep}. 

The SDSD dataset that supports the findings of this study is available in Google Drive at \url{https://drive.google.com/drive/folders/1-fQGjzNcyVcBjo_3Us0yM5jDu0CKXXrV}, reference number~\cite{wang2021seeing}.

The MIT Adobe FiveK dataset that supports the findings of this study is available in MIT CSAIL at \url{https://data.csail.mit.edu/graphics/fivek/fivek_dataset.tar}, reference number~\cite{wang2021seeing}. 

These data were derived from the following resources available in the public domain: 
\begin{itemize}
    \item LOL dataset: \url{https://drive.google.com/file/d/157bjO1_cFuSd0HWDUuAmcHRJDVyWpOxB/view}
    \item SDSD dataset: \url{https://drive.google.com/drive/folders/1-fQGjzNcyVcBjo_3Us0yM5jDu0CKXXrV}
    \item MIT Adobe FiveK dataset: \url{https://data.csail.mit.edu/graphics/fivek/fivek_dataset.tar}
\end{itemize}
\section*{Conflict of interest statement}
We declare that they have no conflicts of interest.
\section*{Acknowledgements}
This work is supported in part by the Beijing Natural Science Foundation under No. L221013 and the National Natural Science Foundation of China under Grant Nos. 62102162 and 62203184.

\bibliographystyle{eg-alpha-doi}

\bibliography{egbibsample}
\end{document}